\documentclass[11pt,a4paper]{article}
\usepackage[acceptedWithA]{tacl2021v1}
\usepackage{times}
\usepackage{latexsym}
\usepackage{comment}
\usepackage{multirow}
\usepackage{hhline}
\usepackage{graphicx}
\usepackage{hyperref}
\usepackage{natbib}
\usepackage{color,soul}
\usepackage{hhline}

\usepackage[T1]{fontenc}

\usepackage[utf8]{inputenc}

\usepackage{microtype}

\usepackage{inconsolata}

\usepackage{graphicx}

\usepackage{booktabs}
\usepackage{float}
\usepackage{tabularx}
\usepackage{todonotes}
\usepackage{threeparttable}
\usepackage{blindtext}
\usepackage[dvipsnames]{xcolor}
\usepackage{algorithm}
\usepackage{algpseudocode}
\usepackage{amsmath}
\usepackage{makecell}
\usepackage[font=small,skip=1pt]{caption}

\usepackage{tacl2021v1}

\usepackage{xspace,mfirstuc,tabulary}

\newif\iftaclinstructions
\taclinstructionsfalse 
\iftaclinstructions
\renewcommand{\confidential}{}
\renewcommand{\anonsubtext}{(No author info supplied here, for consistency with
TACL-submission anonymization requirements)}
\newcommand{\instr}
\fi

\iftaclpubformat 

\else

\fi

\title{CorefInst: Leveraging  LLMs for Multilingual Coreference Resolution}

\author{
   Tuğba Pamay Arslan$^\diamond$ 
   \and
   Emircan Erol$^\diamond$ 
   \and
   Gülşen Eryiğit$^\diamond$ 
   \\
   \ \\
   $^\diamond$İTÜ NLP Research Group
   \\
   Department of Artificial Intelligence and Data Engineering
   \\
   İstanbul Technical University
   \\
   \texttt{[pamay, erole20, gulsen.cebiroglu]@itu.edu.tr}
   \\
   \ \\
 }

\date{}

\renewcommand\hl[1]{#1}

\begin{document}
\maketitle

\begin{abstract}
Coreference Resolution (CR) is a crucial yet challenging task in natural language understanding, often constrained by task-specific architectures and encoder-based language models that demand extensive training and lack adaptability.
\hl{This study introduces the first multilingual CR methodology which leverages decoder-only LLMs to handle both overt and zero mentions}.
The article explores how to model the CR task for LLMs via five different instruction sets using a controlled inference method. 
The approach is evaluated across three  LLMs; Llama 3.1, Gemma~2, and Mistral 0.3.
The results indicate that LLMs, when instruction-tuned with a suitable instruction set, can surpass state-of-the-art task-specific architectures. Specifically, our best model, a fully fine-tuned Llama 3.1 for multilingual CR, outperforms the leading  multilingual CR model (i.e., Corpipe 24 single stage variant) by 2 pp on average across all languages in the CorefUD v1.2 dataset collection.

\end{abstract}
\section{Introduction}

Coreference Resolution (CR) is a semantic-level task in natural language processing (NLP) that involves detecting and clustering mentions referring to the same entity within a text.
An end-to-end CR system consists of two phases: mention detection and coreference linking. In mention detection, all possible referential mentions are extracted from the text. In the next step, mentions referring to the same entity are collected under the same cluster, by resolving which mentions are coreferential.
This task is crucial for various higher-level NLP applications, including machine translation~\citep{mtCoref}, text summarization~\citep{summarizationCoref}, and question answering systems~\citep{qaCoref}, as it helps in understanding the coherence and context of the text.

\begin{figure}[t!]
    \centering    \includegraphics[width=0.45\textwidth]{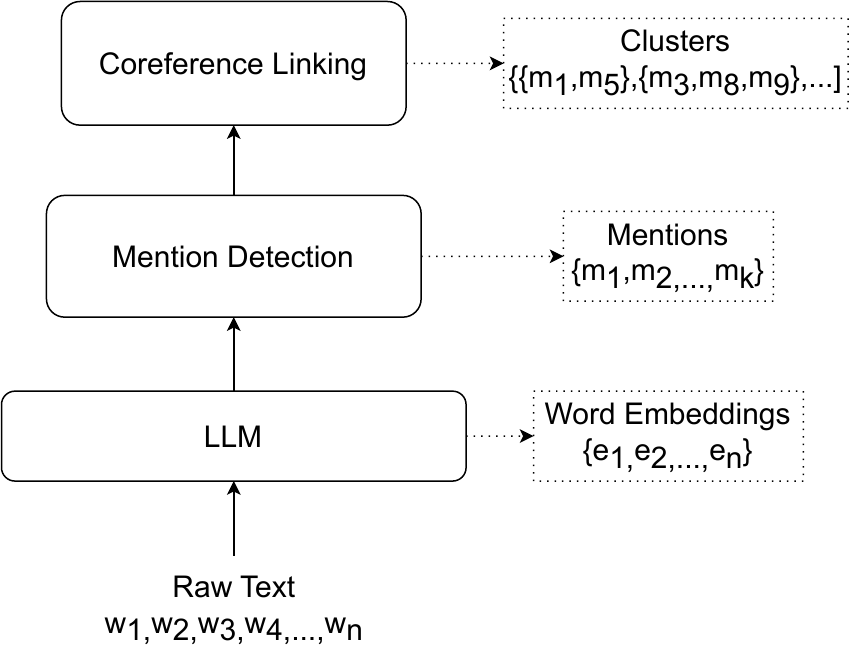}
    \caption{A sample flow of a traditional CR-specific neural architecture.}
    \label{fig:existingArchitecture}
\end{figure}

Large Language Models (LLMs) are traditionally classified into three categories: encoder-only models (e.g., BERT), encoder-decoder models (e.g., T5), and decoder-only models (e.g. GPT, Llama). Initial task-specific neural CR  models~\citep{lee-etal-2017-end, straka-2023-ufal} incorporate an encoder-only LLM at the bottom layer to generate word embeddings, followed by a specialized neural architecture where the LLM layer is also fine-tuned during the training process, illustrated in Figure~\ref{fig:existingArchitecture}.
However, task-specific models require specialized architectures, vast amount of training data, and an extensive fine-tuning/training process from scratch, leading to limited generalizability and flexibility across new datasets, languages, tasks, and domains. 
Furthermore, due to their niche nature, these models cannot exploit decoder-based LLMs directly, which have been rapidly developing. On the other hand, instruction tuning has emerged as a powerful paradigm, enabling decoder-only LLMs to adapt to diverse tasks through carefully crafted prompts or instructions, thereby addressing these limitations. 
It is important to note that recently, LLMs are often used interchangeably with decoder-only models in the literature, particularly due to the popularity of autoregressive models like GPT. From this point onward in the article, we will refer to decoder-only LLMs simply as LLMs.

\begin{figure}[h]
    \centering
    \begin{tabular}{llllll}
         \textcolor{brown}{(on)} & Odešel, & ale & \textcolor{brown}{(on)} & vrátil & se.\\
          \textcolor{brown}{He} & left & but & \textcolor{brown}{he} & came & back.
    \end{tabular}\\
    \smallskip
    \textit{He left but came back.}
    \caption{Zero mention in pro-drop language example}
    \label{fig:prodropExample}
\end{figure}

Despite notable progress in CR, the technologies developed to tackle challenges such as handling zero mentions and generalizing across various languages \hl{are still evolving
}~\cite{crac2024,PAMAYARSLAN2025101681}.
Recent CR studies \citep{chen-etal-2021-tackling, straka-2023-ufal, PAMAYARSLAN2025101681} address zero mentions and their coreferential relations alongside overt mentions. Zero mentions are not explicitly defined in an original sentence, and result from dropped-pronouns in pro-dropped languages (e.g., Chinese, Hungarians, Turkish).
Figure~\ref{fig:prodropExample} presents a sample Czech sentence containing zero mentions along with its English translation. 
As seen, the zero pronoun, `on' (Eng. he), is not explicitly stated in the Czech sentence; in other words, it is omitted. CR systems considering zeros also incorporate these latent coreferential relations arising from dropped pronouns.

To explore the effectiveness and potential of LLMs in multilingual CR, this study seeks answers to the following main research questions:
\begin{itemize}
    \item How can LLMs be effectively leveraged for multilingual CR, including the resolution of both overt and zero mentions?
    \item Can instruction-tuned LLMs outperform traditional, task-specific architectures for multilingual CR? 
    \item How well instruction-tuned LLMs resolve coreferences arising from zero mentions?
\end{itemize}

To address these questions  three state-of-the-art LLMs —Llama 3.1, Gemma-2, and Mistral 0.3— were initially analyzed using five different instruction sets to provide preliminary insights into their CR capabilities. Based on these initial evaluations, the most effective LLM-instruction set pair was selected for further analysis. The selected model, hereafter referred to as CorefInst, 
was first examined under full training and few-shot settings, and then
evaluated in an end-to-end manner on predicted mentions (both overt and zero) and compared against CorPipe24-single~\cite{straka2024corpipe} 
on the CorefUD  dataset v1.2~\citep{corefud12}, yielding better performances. Finally, the capability of CorefInst\footnote{The source code is available at \href{https://github.com/itunlplab/coref-inst}{this URL}.} to resolve zero mentions was specifically assessed, demonstrating superior performance on pro-drop languages.

The major contributions of the article are: 1) the first instruction-tuned LLM for multilingual CR, and 2) a controlled inference method to model the CR task for LLMs. The proposed methodology for instruction-based fine-tuning of LLMs as a coreference resolver also incorporates zero mentions in addition to overt ones.
The article is structured as follows: Section~\ref{sec:prework} introduces the available studies in the literature, Section~\ref{sec:proposed_model} presents our proposed model in detail, Section~\ref{sec:experimental_setup} gives the experimental settings and results, and 
Section~\ref{sec:conclusion} concludes the article. 




\section{Related Work}\label{sec:prework}

Early neural CR studies typically employ task-specific neural architectures which utilize an encoder component of a large language model (i.e., encoder-only or encoder-decoder LLMs) to generate word embeddings. Initially, these approaches exclusively focus on the coreference linking stage~\citep{clark-manning-2016-improving,
wiseman-etal-2016-learning}, 
but in recent years, end-to-end methods, where mention detection and coreference linking stages are trained either jointly or sequentially, have become 
as a dominant approach in the CR literature~\citep{lee-etal-2018-higher, kantor2019coreference, joshi2020spanbert, cawCoref, straka2024corpipe, mscawCoref}.

With the increasing popularity of generative LLMs which contain a decoder component (e.g., T0~\cite{t0llm}, Llama~\cite{llama}, GPT~\cite{gpt}), adapting them to NLP tasks using zero-/few-shot prompting or learning techniques has recently emerged as a prominent research strategy. However, CR studies in this context are still in their early stages.
~\citet{brown2020language} demonstrated that prompting GPT-3 can resolve coreferential relations. 
\citet{yang-etal-2022-gpt} proposes a question-answering-based prompt-engineering approach to assess the capabilities and limitations of generative, pre-trained GPT-variant LLMs in the CR, demonstrating that these models perform poorly on the task without fine-tuning.
\citet{t0llm} introduces T0, a zero-shot generalization of T5~\cite{t5llm}, which transforms various supervised datasets into task-specific prompts, including those for CR.
\citet{gan2024assessing} evaluates various LLMs for their CR abilities using zero- and few-shot prompting, concluding that LLMs still have potential for improvement in their CR performances.
\citet{le2024are} presents a prompt-based CR system by employing instruction-tuned LLMs (e.g., InstructGPT, Llama-2-Chat). Besides prompt-based approaches, sequence-to-sequence CR models are also seen in the CR literature~\citep{zhang-etal-2023-seq2seq, liu2022autoregressive, bohnet-etal-2023-coreference, crac24_text2textCR}.

The CR studies mentioned up to this point have mainly been developed for English, largely due to the availability of standardized annotated dataset~\cite{ontonotes}. \hl{However, this dataset is not publicly available due to licensing restrictions.}. Meanwhile,
multilingual CR studies~\citep{adaptCRAC, pamaycrac, straka2024corpipe} are advancing rapidly in this field, particularly due to the CorefUD initiative and shared tasks conducted in recent years~\citep{crac2023, crac2024}. 
CorefUD~\citep{corefudGeneralReference} is an initiative to collect coreference corpora in various languages and harmonize them to the same scheme and data format (CoNLL-U)~\citep{UD2021}. 
The latest version, CorefUD v1.2~\citep{corefud12}, consists of 21 datasets for 15 languages, such as English, Russian, French, Norwegian and so on. Additionally, the CorefUD, which also includes pro-drop languages (e.g., Turkish and Catalan) has facilitated the development of CR literature for these languages.




\section{The Proposed Method}\label{sec:proposed_model}
This study proposes a methodology for instruction-based fine-tuning of an LLM for the multilingual CR task. 
This section presents the entire instruction engineering process, all the necessary steps for data processing and mention clustering, and the proposed controlled inference method.
One should note that the instruction and input/output format selection are inter-dependent steps. That is, a recursive search combined with a feedback loop method is employed to refine and finalize their formats.



\subsection{Instruction Engineering}

The proposed method accepts a tuple as an input consisting of three parts: <an instruction, masked input sentences, output sentences>, exemplified in Figure~\ref{table:IOexample}.
Basically, the model learns to generate the expected output by determining which coreferences are present in masked input sentences, taking the given instruction into consideration. An instruction is a directive that provides the necessary preliminary information about a task.
This part may contain details about the input, and the expected output format, and also include some restrictions/limitations that should be considered by the LLM when generating outputs.

We explore five different instruction sets, provided in Table~\ref{table:allInstructions} in the Appendix section. Each instruction set has its own characteristic, and may encompass a range of information, including language in focus, different task definitions, explanations, and errors identified in the model's generated outputs.
For instance, the first instruction set (Inst.\#1) is the most detailed one, presenting straightforward statements regarding the task, input/output format, and expectations from the LLM as a coreference resolver. The second instruction set (Inst.\#2) employs an alternative task definition by also providing a sample input/output pair.
The remaining instruction sets represent various modifications of each other. Notably, the third (Inst.\#3)  and fifth (Inst.\#5) instruction sets incorporate a non-common keyword in CR, `coherent'\footnote{\hl{The term `coherent' was proposed by ChatGPT during our preliminary experiments aimed at probing the extent of the LLM’s knowledge regarding the CR. This observation implies that the LLM may implicitly associate these two concepts within its internal representational framework.}}, along with `coreferential'.
\hl{Although `coherent' is not a commonly used term in CR literature, in this study it is used to encourage the model to consider discourse-level consistency, as coreference resolution is closely linked to textual coherence and is often used as one of its measurable indicators}~\citep{coherentsupport}. \hl{It should be noted that this usage is experimental and serves as a complementary guide rather than a replacement for traditional terminology.}
Additionally, the fourth (Inst.\#4) and fifth instruction sets provide a semantically related but non-coreferent mention pair example to illustrate \hl{whether a coreferential relation exists }(e.g., ``author'' and ``book'' represent different entities), with Inst.\#4 containing only the term `coreferential'. One should note that the keyword `coherent' and this additional example are included in instruction sets (i.e., \#3, \#4, and \#5) based on the models' generated errors observed during the instruction engineering process. Moreover, each explored instruction set is complemented by an instruction directing the LLM to identify coreferential relations of zero mentions (depicted in italic font in Table~\ref{table:allInstructions}) as well.


\begin{table*}[th]
\begin{threeparttable}
    \centering
    \begin{tabular}{|p{0.98\textwidth}|}
    \hline
    \textbf{Instruction}\\
    \hline
    {For every mention in <m> </m> tags examine if there is any coreferential/coherent mention.\newline
If two mentions represent the same entity write the same number instead of MASK after closing mention tag </m>.\newline
For example: author and book represent different entities.\newline
Do not change anything else other than MASK.\newline
\textit{Where you see </z>@ there is a zero mention, which is normally not written but you also need to link them with other mentions.\tnote{*}}}\\ 

    \hline
    \textbf{Input} \\
    \hline
    {<m> Leges-legkedvesebb napom </m>\#\textcolor{NavyBlue}{MASK} az volt, <m> mikor </m>\#\textcolor{NavyBlue}{MASK} tavaly június 6.-án mentem a <m> T. sziklási Sport Klub </m>\#\textcolor{NavyBlue}{MASK} nagyobb tagjaival </z>@\textcolor{NavyBlue}{MASK} és <m> Pista bácsival </m>\#\textcolor{NavyBlue}{MASK} <m> az országos váltóbajnokságra </m>\#\textcolor{NavyBlue}{MASK} . [MID] Nagyon örültem, mikor <m> Pista bácsi </m>\#\textcolor{NavyBlue}{MASK} mondta , \_ hogy megyek <m> a versenyre </m>\#\textcolor{NavyBlue}{MASK} , attól a naptól fogva rendszeresen edzettem otthon is és az edzésen is.}\\
    \hline
    \textbf{Output} \\
    \hline
    {<m> Leges-legkedvesebb napom </m>\#\textcolor{OrangeRed}{0} az volt, <m> mikor </m>\#\textcolor{OrangeRed}{0} tavaly június 6.-án mentem a <m> T. sziklási Sport Klub </m>\#\textcolor{OrangeRed}{1} nagyobb tagjaival </z>@\textcolor{OrangeRed}{1} és <m> Pista bácsival <m>\#\textcolor{OrangeRed}{2} <m> az országos váltóbajnokságra </m>\#\textcolor{OrangeRed}{3} . [MID] Nagyon örültem, mikor <m> Pista bácsi </m>\#\textcolor{OrangeRed}{2} mondta, \_ hogy megyek <m> a versenyre </m>\#\textcolor{OrangeRed}{3} , attól a naptól fogva rendszeresen edzettem otthon is és az edzésen is.}\\
    \hline
    \end{tabular}
    \begin{tablenotes}
        \item {(Eng.) <m> My favorite day </m>\#\textcolor{OrangeRed}{0} was <m> when </m>\#\textcolor{OrangeRed}{0} I went to <m> the national relay championship </m>\#\textcolor{OrangeRed}{3} on June 6 last year with the older members of the <m> T. sziklási Sport Club </m>\#\textcolor{OrangeRed}{1} and <m> uncle Pista </m>\#\textcolor{OrangeRed}{2}. [MID] I was very happy when <m> uncle Pista </m>\#\textcolor{OrangeRed}{2} told me that I was going to <m> the competition </m>\#\textcolor{OrangeRed}{3}, from that day on I trained regularly both at home and in training.} 
    \end{tablenotes}
  
    \captionof{figure}{A sample tuple containing Instruction\#5 (see Table~\ref{table:allInstructions} for all explored instructions) with a masked input in Hungarian, and with a corresponding output. The English translation of the output sentence is also provided at the bottom.}
    \label{table:IOexample}
    \end{threeparttable}
\end{table*}

Table~\ref{table:IOexample} presents a sample tuple containing Inst\#5 with a masked input in Hungarian, and with a corresponding output. The first sentence of the instruction set explains how to resolve coreferential relations for each mention, which is encapsulated by specified mention tags.
When the instruction contains only the `coreferential' keyword, it is observed that the model tends to over-merge clusters (i.e., collecting many non-coreferential mentions into the same cluster).  \hl{Consequently, an additional keyword (i.e., coherent) is used together with ‘coreferential’ to ensure that the model pays attention to referential links that depend more on the flow of the discourse rather than on sentence structure or word repetition.}
In the second sentence, the model learns to assign a predicted cluster number solely to the \texttt{MASK} token of each mention.
The third line provides the above-mentioned example.
The forth statement in the instruction set enables to guide the LLM to generate outputs that are as close as possible to expected results, that is, a generated output with only updated \texttt{MASK} token with a predicted cluster number. The last one, written in italic font, is an extra explanation to be used for coreference linking for only pro-dropped languages.
During the instruction engineering process, the performance of each different LLM-instruction set pair is examined, and those with the highest achievements are chosen for further experiments.
These initial analyses help identify high-performing pairs, offering insights into the potential performance of the proposed model using full training and few-shot learning on the entire dataset. 

\subsection{Data Processing and Mention Clustering}

The proposed model is trained and evaluated on CorefUD v1.2 dataset~\citep{corefud12} collection, which are formatted in CoNLL-U~\citep{UD2021}, which is a standardized column-wise format used for annotating and representing linguistic information at the morphological, syntactic, and semantic levels in NLP. Therefore, a data processing phase is required to make these datasets compatible with the instruction-based fine-tuning process of an LLM.
This section presents the pre-processing step on the raw documents to prepare the masked input sentences, and the post-processing step on the generated outputs to re-convert them to CoNLL-U format for evaluation purposes.
\begin{figure*}[t!]
    \centering
    \includegraphics[width=0.98\textwidth]{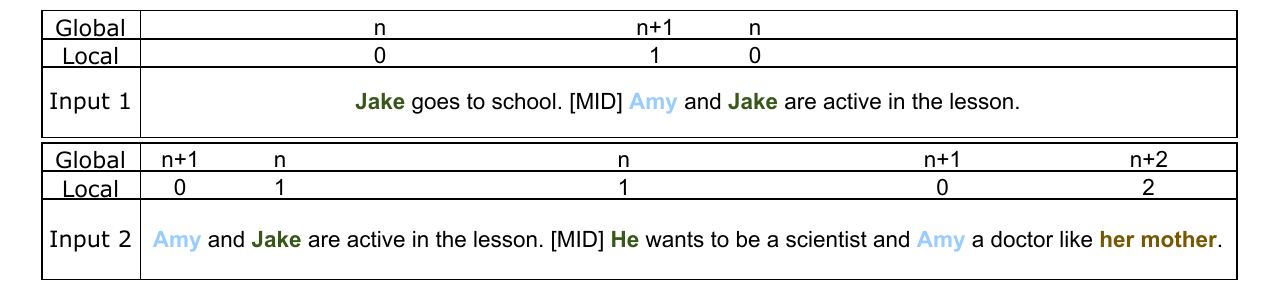}
    \caption{Merging clusters with overlapping frames}
    \label{fig:overlap}
\end{figure*}
\interfootnotelinepenalty=10000

The masked input sentence is a pre-processed version of the original sentence, incorporating mention boundary markers (i.e., \texttt{<m>,</m>} for overt mentions, and \texttt{</z>} for zero mentions) and coreference relation placeholders (i.e., \texttt{MASK}), as presented in the second part of Figure~\ref{table:IOexample}, named `Input'.
Mentions may be considered as word sequences which are surrounded by \texttt{<m>}, and \texttt{</m>} tags for the initial and final tokens of an overt mention, respectively. Similarly, \texttt{</z>} is inserted at the exact location of a dropped pronoun in a sentence to indicate zero mentions. One should note that the length of zero mentions is always 1\footnote{Syntactically, zero mentions are represented as a single node in the dependency tree of a sentence in which they exist, as they are considered to consist of only a single token.}, thus, only closing mark is used for these type of mentions. The main aim of the representation\footnote{This representation is inspired by existing studies in the CR literature~\citep{wu-etal-2020-corefqa, zhang-etal-2023-seq2seq} as well as our initial experiments on various representations.} should be less complex, easily interpretable by the model and human-readable. Therefore, we use only three boundaries (<m>, </m>, <z>) without any numbers for entity clusters or mentions to reduce complexity of the representation, and consequently, difficulty in post-processing.
In addition, the \texttt{MASK} token is used with \# and @ symbols to indicate overt and zero mentions, respectively. This helps to distinguish mention types during the post-processing stage (i.e., converting the predicted output of the LLM to the CoNLL-U data format).

As the final stage of the pre-processing, frames are created from the masked input sentences with a pre-defined length of sequences. The final masked input for the LLM is formed by concatenating two consecutive frames with a separator (i.e., [MID]), which helps to preserve the coreferential relation chain among generated outputs for a document. It should be noted that the length of a document, in terms of the number of tokens, may exceed memory limits or maximum segment length of LLMs. The primary motivation behind using frames is to enable the model to better handle long documents by breaking them down into smaller, more manageable segments (called frames).

The post-processor converts the model's generated output into the CoNLL-U format by transferring the predicted cluster numbers into raw documents (i.e., those having no coreference annotations). The flow of this conversion is illustrated in Figure~\ref{fig:overlap}. For simplicity, each frame consists of only one sentence, meaning the proposed model makes prediction on two sentences in a masked input. The figure provides two different inputs (i.e., Input 1 and Input 2) containing overlapping frames: the right-side (i.e., the second after [MID] separator) of the Input 1 is exactly the same as the left-side of the Input 2. The model predicts a cluster number for each mention, which is indicated just above the mention in the table as variable `local'.
Local number is a predicted cluster number generated as the model's output, and is only meaningful within the frames of a single input. 
However, a global set of cluster numbers is required to collect all coreferential mentions in the same cluster within a document. Therefore, during the post-processing, each mention is also assigned to its global cluster number with the help of overlapping frames.
For example, while the first and second occurrences of `Jake' in the first input have the same local cluster number of 0, the mention `Amy' is assigned to a different local cluster number of 1. Since this is the first input, each local cluster is directly mapped to a newly created global cluster (i.e., 0$\rightarrow$n, and 1$\rightarrow$n+1) and the mappings between them are kept for further inputs. Then, the second input is processed. 
In the second input, there are two occurrences of `Amy' in the first and second frames. Since the global cluster numbers for the mentions in the first frame are already established, these occurrences of `Amy' are directly assigned to an available global cluster, n+1. Similarly, since the mention `He' is coreferential with `Jake', its global cluster number is also assigned to n. In the second frame, there is a new mention `her mother'. Since this is a newly referenced mention among all inputs, a new global cluster is created for this mention with number n+2.
In this way, all inputs are post-processed sequentially while preserving the coreferential relation chains within a document. The pseudo-code of this process is explained in Algorithm~\ref{alg:cluster}.

\begin{algorithm}
\caption{Merge Clusters over Frames}
\label{alg:cluster}
\begin{algorithmic}[1]
\State \texttt{globCls} is initialized.
\ForAll{$before, after \in frameTuples$}
    \State $map = \{\}$
    \For{$mPos, localNo \gets before$}
        \State $map[localNo] \gets globCls[mPos]$
    \EndFor
    \ForAll{$mPos, localNo \in after$}
        \If{$localNo \not \in map$}
            \State $i\gets \max(globCls) + 1$\
            \State $map[localNo] \gets i$
        \EndIf
        \State $globCls[mPos] \gets map[localNo]$
    \EndFor
\EndFor
\end{algorithmic}
\end{algorithm}


Initially, a global cluster dictionary, \texttt{globCls}, is initialized by the first frame of the first input along with the model's predicted cluster numbers for that corresponding frame (referred to as `local' in Figure~\ref{fig:overlap}).
Keys of \texttt{globCls}, \texttt{mPos}, represent a mention's position within a document (i.e., start and end words' indices).
The following input-output pairs are then processed sequentially in the order they appear within a document.
\texttt{before} and \texttt{after} are lists of mentions that exist in the first and second frames of an input, respectively. The local cluster numbers and word indices of these mentions are stored as attributes of each mention in these lists.
The overlapping structure of consecutive inputs (Figure~\ref{fig:overlap}), enables the construction of a mapping dictionary (i.e., \texttt{map}), between the
local and global cluster numbers across consecutive frames (i.e., second frame of the k\textsuperscript{th} input and first frame of the (k+1)\textsuperscript{th} input). 
For each input (e.g., frame tuples), global cluster numbers that are established for the first frame (\texttt{before}) are stored into the mapping dictionary. Then, for the second frame, if any local cluster number has not yet been assigned to a global cluster number, then a new cluster is created with number $i$, and the mention is inserted to this newly created cluster. At the end, mentions from the second frame, are inserted into a global cluster.






\subsection{Controlled Inference Method}

After an LLM is fine-tuned using the aforementioned tuples (i.e., <instruction, masked input, output>), it learns and adapts to the coreference linking process.
Specifically, the model becomes capable of linking overt and zero mentions with the assistance of the developed data processors.

The coreferential relations of each mention are resolved sequentially as they appear in an input sentence during both training and inference. Due to the causal nature of LLMs, these models correlate information in an autoregressive manner for each decision, starting from the current mention and moving towards the descendant mentions and their predicted cluster numbers. Consequently, we developed a controlled inference methodology to mitigate the model's tendency to make random predictions when formulating new decisions during the inference phase, and to ensure that it relies on prior context for its decision-making. This approach focuses solely on predicting the cluster number to replace the \texttt{MASK} token in the input, while automatically  keeping all other tokens the same in the original sentence.



\begin{figure}[ht]
    \centering    \includegraphics[width=\linewidth]{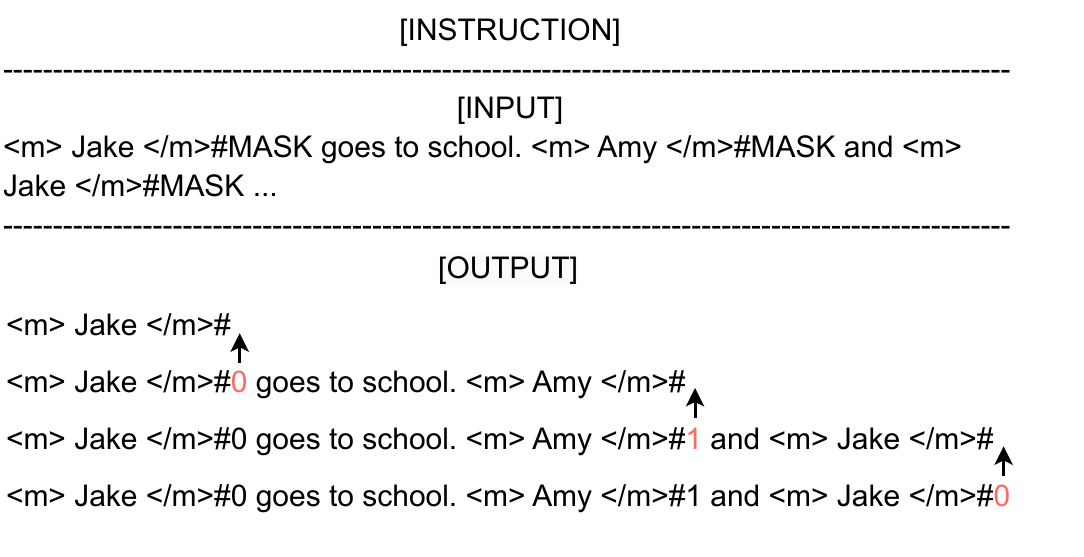}
    \caption{Inference steps}
    \label{fig:genExample}
\end{figure}

The inference steps for English samples are illustrated in the [OUTPUT] section of Figure~\ref{fig:genExample}, which also includes the masked input sentences under the [INPUT] section.
Each inference step is presented on a new line in the [OUTPUT] section. The arrow shows the position of \texttt{MASK} tokens where the model is called for inference. 
The model predicts the current \texttt{MASK} token by considering all preceding context up to the start of the output. The non-colored parts, namely up to \texttt{MASK}, are given to the model and then the red part is generated by the model during the inference.
For example, for the first \texttt{MASK} token, the model generates a cluster number 0, immediately after the position indicated by the arrow, by replacing the current \texttt{MASK} token with the predicted cluster number.
In each inference step, the pre-context is updated with a newly predicted cluster number to maintain the flow of cluster information in subsequent inference steps. The arrow then moves to the next \texttt{MASK} token, reaching the position where the model will be called again. In this way, all \texttt{MASK} tokens are predicted sequentially, allowing the model to utilize its previous decisions for the current decision.

This inference mechanism is termed the controlled inference strategy which provides several advantages: 1) It significantly reduces computational load during inference by calling the model on only at specified points, leading to faster processing times, 2) It maintains the original text structure, ensuring consistency in input/output formats and enabling easy conversion to the desired data formats. By focusing the model's efforts exclusively on predicting cluster number, the proposed framework achieves a balance between accuracy, speed, and practical applicability.
\section{Experimental Setup \& Results}\label{sec:experimental_setup}

This section introduces the details of the experimental setup for the fine-tuning process, as well as the performances of the tested models.

\subsection{Experimental Setup}

This study validates the proposed methodology on instruction-tuned versions of three recent LLMs: Llama 3.1\footnote{\label{llamaLink}\url{/unsloth/Meta-Llama-3.1-8B-Instruct-bnb-4bit}}, Gemma 2\footnote{\url{/unsloth/gemma-2-9b-it-bnb-4bit}}, and Mistral 0.3\footnote{\url{/unsloth/mistral-7b-instruct-v0.3-bnb-4bit}}. These models are fine-tuned using instruction-based method specifically for the coreference linking phase in CR.
Low-Rank Adaptation (LoRA)~\cite{hu2021loralowrankadaptationlarge} with a rank of 16 (R=16), and 4-byte quantization is used during training (i.e., few-shot and full) to use resources efficiently.
This setup is designed to balance computational efficiency with model performance, leveraging the capabilities of the LLM while maintaining tractable training times and resource requirements. 
In this study, full training denotes training the LLM on all samples within a dataset, whereas few-shot (specifically 5-shot) training involves training using only 5 examples from each dataset.
For the initial experiments, training and validation splits are chosen from a subset of an English dataset (en\_gum) from CorefUD v1.2. The LLMs are then fully trained with 3 different seeds (i.e., 3 models for each LLM-instruction set pair in total) and subsequently evaluated for each instruction set using these data splits. The average performance across these runs is presented as the final performance for each LLM-instruction set pair. This approach is essential to enhance the reliability of our preliminary results.
The full training procedure is then conducted on the entire CorefUD v1.2 using the highest-performing LLM-instruction set pair, which is accepted as the proposed model in this study, CorefInst. All models are trained and evaluated on a single NVIDIA V100 (32 GB) or A100 (80 GB) GPU.

The CorefUD v1.2 dataset collection~\citep{corefud12} is utilized for the training and evaluation of the models under investigation. This dataset is multilingual and comprises documents formatted in CoNLL-U, along with coreference annotations.
The CorefUD v1.2 contains 21 datasets from 15 languages, also containing pro-dropped languages with zero-mention annotations (see Table~\ref{table:corefud-datasets-stats} for the dataset statistics).

The initial experiments on a subset of English documents take $\sim$20 minutes on a V100 GPU (batch size: 4, epochs: 5). The 5-shot training takes 10 minutes (batch size: 16, epochs: 5), while full training on CorefUD v1.2 takes $\sim$15 hours on an A100 GPU (batch size: 16, epochs: 3). The pre-defined length of each frame is up to 1600 and the total of an input tuple is up to 7168 for the proposed model.


CorPipe24~\cite{straka2024corpipe}, the most advanced version of the CorPipe system family and the winner of the latest multilingual CR shared task~\cite{crac2024}, is used as the leading multilingual CR model for comparing our proposed model in an end-to-end manner. We used its single-stage variant, which employs a single pre-trained encoder model to jointly predict empty nodes (i.e., zero mentions), coreferential overt mentions, and coreferential links between them. This variant is the large-sized, single-stage model checkpoint that achieves the highest development performance across all CorefUD v1.2 datasets. Since the proposed model in this study was also selected using the same approach, comparing it with the single-variant of the CorPipe24\footnote{~\citet{straka2024corpipe} also introduces several variants employing different ensembling strategies; however, ensemble models fall beyond the scope of this article.} system provides a fairer comparison. For end-to-end evaluation of CorefInst (i.e., on predicted overt mentions and zeros), the mention/zero detection output from CorPipe24 is directly used, ensuring both models operate on the same predicted input during the coreference linking stage.
\hl{Similar to the initial experiments, during the end-to-end evaluation of our proposed model, CorefInst, and its comparison with CorPipe24, all models were trained and evaluated using 3 different random seeds.
The final performances reported for each model are presented as the average of these 3 runs. Moreover, we apply statistical significance testing (SST) to the performance differences between these models\footnote{\hl{Due to the resource-intensive nature of such experiments — in terms of time, computational cost, and carbon footprint} \citep{deepSignificance} — \hl{we refrain from applying SST across all intermediate stages and analyses. Instead, we apply it only to our end-to-end experiments on predicted mentions.}} to show the robustness of the performance both in terms of the average and across the datasets in each language. We apply the Almost Stochastic Order (ASO) test with confidence level $\alpha = 0.95$}~\citep{ deepSignificance} \hl{on multiple runs (i.e., 3 runs) with random seeds.}

The best model is evaluated on the development splits of the CorefUD v1.2 since test splits are blind. The average CoNLL coreference metric which is the average of three different metrics: MUC, B-Cubed, and entity-based CEAF are used as the main performance metric.
One should note that, these CR metrics (MUC, B$^3$, CEAF$_e$, and CoNLL—the average of the three) evaluate the quality of mention clustering (i.e., entitites) from different perspectives, categorized as link-based, mention-based, and entity-based metrics, respectively. None of these metrics were specifically designed to evaluate the coreference performance of dropped pronouns. However, with the CorefUD initiative, CR studies on pro-drop languages have gained greater significance. In this context, the anaphor-decomposable score (i.e., zero metric) has been recently proposed~\cite{crac2022} as a dedicated metric for evaluating a model on coreferences arising from zero mentions. As a final evaluation, the performance of the best model is also presented with this metric. An available CR scorer\footnote{\url{https://github.com/ufal/corefud-scorer}} is used for the evaluation to be sure that the performances are compatible with the literature.

\subsection{Results \& Discussion}

The results of the initial experiments on a subset of English dataset is provided in Table~\ref{table:resultsInitExperiments}, where 
each inner cell shows the average performance of 3 different runs (with also standard deviation) for each LLM-instruction set pair with different seeds, and the last row reports the average performances of the LLMs across instruction sets, along with the standard deviation.
The best performance of Llama is obtained by Inst\#5 (76.15\% CoNLL), whereas Gemma and Mistral reach their highest performances with Inst\#3 and Inst\#1 (73.82\% and 72.46\%, respectively).
When the overall table is examined, the lowest average performance is obtained with Mistral, followed by Gemma. Llama achieves the highest performance for all instructions as well as on average across instructions. Based on these initial results, it has been decided to continue with the \hl{highest scoring} LLM-instruction pair (Llama-Inst\#5) for further analysis.

\begin{table}[H]
\centering
\resizebox{\columnwidth}{!}{  
\begin{tabular}{l|c|c|c|}
&  Llama 3.1 & Gemma 2 & Mistral 0.3 \\ \hline
\multicolumn{1}{l|}{Inst.\#1} & 75.57\tiny{$\pm$0.49} & 72.45\tiny{$\pm$1.71} & \textbf{72.46\tiny{$\pm$1.37}}         \\ 
\multicolumn{1}{l|}{Inst.\#2} & 74.71\tiny{$\pm$0.32} & 73.51\tiny{$\pm$0.84} & 71.28\tiny{$\pm$1.64}         \\ 
\multicolumn{1}{l|}{Inst.\#3} & 74.91\tiny{$\pm$1.35} & \textbf{73.82\tiny{$\pm$2.13}} & 71.44\tiny{$\pm$2.59}       \\ 
\multicolumn{1}{l|}{Inst.\#4} & 75.54\tiny{$\pm$1.48} & 73.43\tiny{$\pm$0.96} & 71.58\tiny{$\pm$2.15}        \\ 
\multicolumn{1}{l|}{Inst.\#5} & \textbf{76.15\tiny{$\pm$1.16}} & 73.25\tiny{$\pm$1.07} & 71.55\tiny{$\pm$2.17}          \\ \hline
\multicolumn{1}{l|}{AVG} & \textbf{75.38\tiny{$\pm$0.57}} & 73.29\tiny{$\pm$0.52} & 71.66\tiny{$\pm$0.46}         \\ 
\hline

\end{tabular}
}
\caption{The performances of the initial experiments with a subset of English dataset.}
\label{table:resultsInitExperiments}
\end{table}

\renewcommand{\arraystretch}{2}
\begin{table*}[]
\centering
\addtolength{\tabcolsep}{-0.2em}
\resizebox{\textwidth}{!}{
\begin{tabular}
{llrrrrrrrrrrrrrrrrrrrrrr}
\bottomrule
\vspace{-17mm}
\\
\\
&
System
& \rotatebox[origin=c]{90}{\parbox[c]{1cm}{ \textbf{AVG}}}
& \rotatebox[origin=c]{90}{\parbox[c]{1.75cm}{ ca\_ancora}}
& \rotatebox[origin=c]{90}{\parbox[c]{1.75cm}{ cs\_pcedt}}
& \rotatebox[origin=c]{90}{\parbox[c]{1.75cm}{ cs\_pdt}}
& \rotatebox[origin=c]{90}{\parbox[c]{1.75cm}{ cu\_proiel}}
& \rotatebox[origin=c]{90}{\parbox[c]{1.75cm}{ de\_parcor}}
& \rotatebox[origin=c]{90}{\parbox[c]{1.75cm}{ de\_potsdam}}
& \rotatebox[origin=c]{90}{\parbox[c]{1.75cm}{ en\_gum}}
& \rotatebox[origin=c]{90}{\parbox[c]{1.75cm}{ en\_litbank}}
& \rotatebox[origin=c]{90}{\parbox[c]{1.75cm}{ en\_parcor}}
& \rotatebox[origin=c]{90}{\parbox[c]{1.75cm}{ es\_ancora}}
& \rotatebox[origin=c]{90}{\parbox[c]{1.75cm}{ fr\_democrat}}
& \rotatebox[origin=c]{90}{\parbox[c]{1.75cm}{ grc\_proiel}}
& \rotatebox[origin=c]{90}{\parbox[c]{1.75cm}{ hbo\_ptnk}}
& \rotatebox[origin=c]{90}{\parbox[c]{1.75cm}{ hu\_korkor}}
& \rotatebox[origin=c]{90}{\parbox[c]{1.75cm}{ hu\_szeged}}
& \rotatebox[origin=c]{90}{\parbox[c]{1.75cm}{ lt\_lcc}}
& \rotatebox[origin=c]{90}{\parbox[c]{1.75cm}{ no\_bokmaal}}
& \rotatebox[origin=c]{90}{\parbox[c]{1.75cm}{ no\_nynorsk}}
& \rotatebox[origin=c]{90}{\parbox[c]{1.75cm}{ pl\_pcc}}
& \rotatebox[origin=c]{90}{\parbox[c]{1.75cm}{ ru\_rcr}}
& \rotatebox[origin=c]{90}{\parbox[c]{1.75cm}{ tr\_itcc}}
\\
\\
\vspace{-17mm}
\\
\bottomrule


\multicolumn{1}{l|}
{\multirow{2}{*}{\rotatebox[origin=c]{90}{\parbox[c|]{1cm}{ GOLD}}}}


 &  {CorefInst$_{few}$}
& 73.8 & 77.1 & 74.0 & 70.3 & 56.7 & 81.8 & 76.3 & 69.7 & 76.7 & 83.4 & 76.4 & 67.3 & 72.2 & 65.5 & 77.2 & 82.7 & 93.3 & 67.2 & 64.1 & 67.4 & 84.7 & 66.0 \\


\multicolumn{1}{l|}{} & {CorefInst$_{full}$} & \textbf{85.8} & \textbf{92.0} & \textbf{90.4} & \textbf{87.6} & \textbf{76.1} & \textbf{91.1} & \textbf{85.5} & \textbf{81.8} & \textbf{84.5} & \textbf{95.2} & \textbf{93.0} & \textbf{77.1} & \textbf{85.7} & \textbf{71.1} & \textbf{86.7} & \textbf{91.7} & \textbf{96.7} & \textbf{81.3} & \textbf{82.2} & \textbf{82.3} & \textbf{90.0} & \textbf{80.5} \\

\hhline{========================}

\multicolumn{1}{l|}
{\multirow{2}{*}{\rotatebox[origin=c]{90}{\parbox[c|]{1cm}{ PRED.}}} }

& {CorPipe24}  & 70.5 & 78.9 & 71.0 & 74.3 & 50.3 & \textbf{70.8} & 75.4 & 74.4 & 74.5 & 66.2 & 79.7 & \textbf{69.6} & 62.8 & \textbf{65.5} & 62.2 & 65.9 & \textbf{79.5} & \textbf{80.3} & \textbf{80.1} & 73.4 & 75.4 & 50.1  \\

\multicolumn{1}{l|}{} & {CorefInst$_{full}$} & \textbf{72.3} & \textbf{79.5} & \textbf{71.3} & \textbf{75.8} & \textbf{57.0} & 68.4 & \textbf{77.9} & \textbf{75.5} & \textbf{78.8} & \textbf{68.0} & \textbf{80.8} & 69.4 & \textbf{70.7} & 63.9 & \textbf{64.1} & \textbf{67.1} & 79.1 & 78.9 & 78.4 & \textbf{75.9} & \textbf{75.9} & \textbf{61.6}  \\

\hline
\end{tabular}}
\caption{The performance of the CorefInst on gold and predicted mentions for each individual language. 
}
\label{table:allResultsLlama3.1}
\end{table*}

\renewcommand{\arraystretch}{1.2}
\begin{table*}[]
\centering
\resizebox{.8\textwidth}{!}{
\begin{tabular}
{llrrrrrrrrrrr}
\bottomrule
\vspace{-12mm}
\\
\\
System
& \rotatebox[origin=c]{90}{\parbox[c]{1cm}{ \textbf{AVG}}}
& \rotatebox[origin=c]{90}{\parbox[c]{1.75cm}{ ca\_ancora}}
& \rotatebox[origin=c]{90}{\parbox[c]{1.75cm}{ cs\_pcedt}}
& \rotatebox[origin=c]{90}{\parbox[c]{1.75cm}{ cs\_pdt}}
& \rotatebox[origin=c]{90}{\parbox[c]{1.75cm}{ cu\_proiel}}
& \rotatebox[origin=c]{90}{\parbox[c]{1.75cm}{ es\_ancora}}
& \rotatebox[origin=c]{90}{\parbox[c]{1.75cm}{ grc\_proiel}}
& \rotatebox[origin=c]{90}{\parbox[c]{1.75cm}{ hu\_korkor}}
& \rotatebox[origin=c]{90}{\parbox[c]{1.75cm}{ hu\_szeged}}
& \rotatebox[origin=c]{90}{\parbox[c]{1.75cm}{ pl\_pcc}}
& \rotatebox[origin=c]{90}{\parbox[c]{1.75cm}{ tr\_itcc}}
\\
\\
\vspace{-11mm}
\\
\bottomrule


{CorPipe24} & 68.7 & 77.7 & 56.6 & 70.2 & 60.6 & 79.3 & 75.1 & 67.1 & 63.9 & 77.9 & 58.2 \\




{CorefInst$_{full}$} & \textbf{77.1} & \textbf{86.3} & \textbf{61.5} & \textbf{76.1} & \textbf{71.8} & \textbf{88.0} & \textbf{85.8} & \textbf{70.4} & \textbf{68.7} & \textbf{85.0} & \textbf{77.7}  \\



\hline
\end{tabular}}
\parbox{11cm}{\caption{\label{table:zeroMetrics} Anaphor-decomposable F1 scores* on predicted zero anaphors. 
\\{\scriptsize *These scores are not directly comparable with the CoNLL scores on entire entities in Table~\ref{table:allResultsLlama3.1}.}}}

\end{table*}




The top-performing model (hereafter, CorefInst in Table~\ref{table:allResultsLlama3.1}, and ~\ref{table:zeroMetrics}) is then 5-shot or fully trained and evaluated on CorefUD v1.2.
Table~\ref{table:allResultsLlama3.1} shows the performances of the CorefInst across various configurations.
Firstly, the model's performance is analyzed following fine-tuning using both 5-shot (CorefInst$_{few}$) and full training (CorefInst$_{full}$) strategies on gold mentions. As seen, the CorefInst$_{full}$ outperforms the CorefInst$_{few}$ in each language and their all documents, and also in average by 12 pp across all languages.
The improvements across all datasets range from 3.4 to 19.4, as obtained on Lithunian lt\_lcc) and Antic Slavonic (cu\_proiel). The top five languages where the model performed the best are Lithuanian, Russian (ru\_rcr), Ancient Hebrew (hbo\_ptnk), English (en\_litbank), and Hungarian (hu\_szeged). In addition, in the perspective of pro-dropped languages, which are Catalan (ca\_ancora), Czech (cs\_[pcedt,pdt]), Slavonic, Spanish (es\_ancora), Greek (grc\_proiel), Hungarian (hu\_[korkor,szeged]), Polish (pl\_pcc), and Turkish (tr\_itcc), CorefInst$_{full}$ also outperforms its few-shot counterparts.

The performance of the CorefInst$_{full}$ on the predicted mentions is also analyzed, enabling the end-to-end evaluation of the proposed model, as indicated in the last row of Table~\ref{table:allResultsLlama3.1}. The third line of the table also includes the performance of CorPipe24 for comparison.
\hl{As a reminder, as part of the end-to-end evaluation, the models' performances are reported by the average results across three runs of models.}
\hl{CorefInst$_{full}$ achieves 1.8 pp improvement than CorPipe24 on predicted overt and zero mentions in average (70.5\tiny{$\pm$1.1}\normalsize\% vs. 72.3\tiny{$\pm$1.3}\normalsize\%), which is statistically significant. CorefInst$_{full}$ outperforms CorPipe24 on several languages, with the exception of one dataset in German (de\_parcor) and as well as datasets in French, Ancient Hebrew, Lithuanian and Norwegian.
Among improvements obtained by CorefInst$_{full}$, the increases observed on the Catalan, Czech (cs\_pdt), Old Church Slavonic, German (de\_potsdam), English (en\_gum, en\_litbank), Spanish, Ancient Greek (grc\_proiel), Hungarian (hu\_szeged), Polish (pl\_pcc), and Turkish are statistically significant, whereas the improvements on one dataset each in Czech (cs\_pcedt), English (en\_parcor), and Hungarian (hu\_korkor), as well as on Russian, are not statistically significant. The 5-top significant highest increments are observed on Turkish (11.5 pp),
Ancient Greek (7.9 pp),
Old Church Slavonic (6.7 pp),
English (4.8 pp, en\_litbank), and
German (de\_potsdam) and Polish (2.5 pp for both). 
}


 %


As a final exploration, in Table~\ref{table:zeroMetrics}, \hl{we also report the average performance of three runs of the proposed model (CorefInst$_{full}$) on predicted zero mentions, evaluated across the related datasets of pro-drop languages.} This evaluation indicates how well the model does from the perspective of zero mentions.
In addition to achieving higher performance in the end-to-end evaluation using the CoNLL metric on predicted overt and zero mentions (Table~\ref{table:allResultsLlama3.1}), the proposed model also demonstrates higher coreference linking performance on predicted zeros across \hl{all pro-dropped languages, improving the average performance by 8.4 pp (68.7\tiny{$\pm0.9$}\normalsize\% vs. 77.1\tiny{$\pm1.4$}\normalsize\%), which is statistically significant. On a language-specific basis, all improvements are statistically significant, ranging from 3.3 pp (for Hungarian, hu\_korkor) to 19.5 pp (for Turkish).}
Zero mentions are implicit references to an entity within a sentence, so that resolving their coreferential relations necessitates a comprehensive semantic understanding of the entire text.
This experiment demonstrates that LLMs, pre-trained on diverse and extensive datasets, exhibit greater capability in resolving hidden semantic relations in a text arising from zero mentions, compared to CR-specific neural architectures.

\section{Conclusion}
\label{sec:conclusion}
This article presents the first approach for instruction-based fine-tuning of an LLM for the multilingual coreference resolution (CR) task, uniquely incorporating both overt and zero mentions (i.e., coreferential dropped pronouns). The proposed framework includes comprehensive data processing and mention clustering stages, and a controlled inference method designed to meet the strict data constraints and relation resolution phase. 

 The proposed framework is evaluated across three LLMs, Llama 3.1, Gemma 2, and Mistral 0.3, with five distinct instruction sets. In-depth analysis of the best-performing LLM-instruction set pair (i.e., Llama 3.1 with the fifth instruction) is performed under the settings of few-shot and full training on gold mentions, along with an end-to-end evaluation on predicted mentions including both overt and zero ones. The findings suggest that LLMs, when fine-tuned with an appropriate instruction set, can achieve superior performance compared to state-of-the-art task-specific architectures. Results reveal that the best model (i.e., fully trained Llama 3.1-Inst.\#5, CorefInst$_{full}$) achieves an 12 pp gain in average on gold mentions across 21 documents in 15 languages over its few-shot version. \hl{Furthermore, this model outperforms CorPipe24, the state-of-the-art end-to-end multilingual coreference resolution model on CorefUD v1.2, by an average of 1.8 percentage points which is statistical significant. Moreover, statistically significant language-specific improvements range from 0.6 to 11.5 percentage points.}
\hl{This study also demonstrates that decoder-only models, when properly fine-tuned, can effectively resolve zero anaphors which requires a deep understanding of the contextual focus.}
These findings show the effectiveness of instruction-tuning for flexible, scalable, and high-performing multilingual CR. \hl{At the end, this study provides practical insights to guide future research focused on scaling multilingual CR through general-purpose LLMs instead of CR-specific architectures.}

\section*{Limitations}
\label{sec:limitations}
\vspace{-2mm}
The main limitation of this study is that coreferential relations throughout the document rely on inter-frame resolved coreferential relations within an input tuple. For an entity cluster, the chain of coreferential relations is established via overlapping frames of consecutive input tuples. This may cause that some relations may not be carried further in the document due to a false-negative relation between frames within an input tuple.
In addition, the absence of any mentions in these overlapping frames may also disrupt the coreferential relation chain, leading to the splitting of an entity's mention cluster into sub-clusters, each containing mentions referring to the same entity.

Due to the size and the time and resource complexity required for fine-tuning of the LLMs explored, they could only be fine-tuned with LoRA for the proposed methodology. Therefore, the starting points for the LLMs explored are already their quantized versions.

Studies using LLMs acknowledge the potential limitation arising from the possibility that pretrained LLMs may have been exposed to texts in the evaluation datasets of the tasks in focus during their pretraining phase. This situation might be possible in our study as well; however, the novelty of this paper lies in modeling how this problem can be approached for the first time in the literature.  
Therefore, even if the LLMs were exposed to the evaluation dataset's text, it is unlikely that they have seen it in a way that would be used for instruction-based fine-tuning. Although we didn't provide results in a zero-shot setting, our 5-shot results, which are significantly behind the fully trained model, provide enough clues that pretrained LLMs are not yet capable of solving this complicated CR task.


This study proposed a framework for only coreference linking stage for the CR. The main reason for this is that the mention detection framework planned to be proposed allows the start and end tags for each mention to be placed at any position in the sentence, which requires a long inference time. Developing an effective, accurate, and efficient controlled inference mechanism for this stage is included as future work in this study.

During the revision period of this article, new open-source decoder-only LLMs, such as~\citet{deepseekai2024deepseekllmscalingopensource}, have been released, reflecting the rapid development in the field. Due to the computational cost of extensive evaluations, we prioritized maintaining the scope of our experiments while acknowledging the potential for future work on newer models.
Since the proposed model is not constrained by a task-specific architecture, future studies may extend our findings by integrating and evaluating this approach with newly developed decoder-only LLMs to further investigate their effectiveness in CR.

\hl{Due to the high computational cost of training and evaluating all possible model-instruction combinations on the full dataset, only the top-performing pair from preliminary experiments was selected for further evaluation. Although multiple runs were conducted to ensure the reliability of initial results, we acknowledge that the differences among instructions were relatively small, and it is possible that other instructions may lead to better performance. Exploring different instructions more thoroughly across various languages and LLMs could be a promising area for future work.}

\hl{Other suggested research directions for multilingual CR include investigating the impact of larger context windows on performance, conducting controlled architectural comparisons between decoder-only and encoder-based LLMs, and examining the tradeoffs between English-centric and multilingual models. These studies could provide deeper insights and contribute to advancing multilingual CR beyond the scope of the current work.}


\section*{Acknowledgment}
This work was supported by the Scientific and Technological Research Council of Turkey (TÜBİTAK) under project grant No. 123E079 within the scope of the TÜBİTAK 2515 (European Cooperation in Science and Technology-COST) program.
The computing resources used in this work were provided by the National Center for High Performance Computing of Turkey (UHeM) under grant number 4021342024, and by the İTÜ Artificial Intelligence and Data Science Application and Research Center.
We would also like to thank the anonymous reviewers and the action editor for their valuable feedback and constructive suggestions, which greatly improved the quality of this article.

  
\bibliographystyle{acl_natbib}

\section*{Appendix}
\appendix
\label{sec:appendix}

\begin{table}[htb]
\centering
\renewcommand*{\arraystretch}{1.0}
\resizebox{\textwidth}{!}{ 
\begin{tabular}{|l|l|}
\hline
Inst. \#1 & \begin{tabular}[c]{@{}l@{}}The text is in English.\\ You are a coreference resolver.\\ Rewrite the sentence considering these rules:\\ Mentions are in \textless{}m\textgreater{}...\textless{}/m\textgreater{}\#MASK format.\\ Group the mentions that refer to same real-world entity.\\ If mentions refer to same thing write the same number instead of MASK.\\ If mentions represent different things write another number. \\
\textit{Where you see </z>@ there is a zero mention, which is normally not written but you also need to link them with other mentions.\tnote{*}}
\end{tabular}                                                        \\ \hline
Inst. \#2 & \begin{tabular}[c]{@{}l@{}}Identify instances of coreference where different mentions refer to the same entity.\\ Rewrite the passage, tagging each term with a unique identifier and indicating coreferential relationships.\\ Ensure accuracy and consider context.\\ Fill MASK with unique number for every entity. Be sure that they represent exactly the same entity, not similar.\\ Example output format:\\ ... \textless{}m\textgreater{}Bertrand Russell \textless{}/m\textgreater{}\#MASK is a good author, I love \textless{}m\textgreater{}The History of Western Philosophy \textless{}/m\textgreater{}\#MASK ...\\ ... \textless{}m\textgreater{}Bertrand Russell \textless{}/m\textgreater{}\#0 is good author,  I love \textless{}m\textgreater{}The History of Western Philosophy \textless{}/m\textgreater{}\#1 ... \\
\textit{Where you see </z>@ there is a zero mention, which is normally not written but you also need to link them with other mentions.\tnote{*}}
\end{tabular} \\ \hline
Inst. \#3 & \begin{tabular}[c]{@{}l@{}}For every mention in \textless{}m\textgreater \textless{}/m\textgreater tags examine if there is any \underline{coreferential/coherent} mention.\\ If two mentions represent the same entity write the same number instead of MASK after closing mention tag \textless{}/m\textgreater{}.\\ Do not change anything else other than MASK. \\
\textit{Where you see </z>@ there is a zero mention, which is normally not written but you also need to link them with other mentions.\tnote{*}}
\end{tabular}                               \\ \hline
Inst. \#4 & \begin{tabular}[c]{@{}l@{}}For every mention in \textless{}m\textgreater \textless{}/m\textgreater tags examine if there is any \underline{coreferential} mention.\\ If two mentions represent the same entity write the same number instead of MASK after closing mention tag \textless{}/m\textgreater{}.\\ \underline{For example: author and book represent different entities.}\\ Do not change anything else other than MASK. \\
\textit{Where you see </z>@ there is a zero mention, which is normally not written but you also need to link them with other mentions.\tnote{*}}
\end{tabular}                                               \\ \hline
Inst. \#5 & \begin{tabular}[c]{@{}l@{}}For every mention in \textless{}m\textgreater \textless{}/m\textgreater tags examine if there is any \underline{coreferential/coherent} mention.\\ If two mentions represent the same entity write the same number instead of MASK after closing mention tag \textless{}/m\textgreater{}.\\ \underline{For example: author and book represent different entities.}\\ Do not change anything else other than MASK. \\
\textit{Where you see </z>@ there is a zero mention, which is normally not written but you also need to link them with other mentions.\tnote{*}}
\end{tabular}                     \\ \hline
\end{tabular}
}

\parbox{15cm}{
\caption{\label{table:allInstructions}The list of instruction sets taken into consideration.
}
}
\end{table}

\cleardoublepage


\begin{center}
\begin{minipage}{\textwidth}
\centering
\begin{tabular}{llllll|ll}
\hline
\multicolumn{1}{l}{\multirow{2}{*}{Document}} & \multicolumn{5}{c}{Total number of}     & \multicolumn{2}{|c}{Distribution} \\
\cline{2-8} 
\multicolumn{1}{c}{}                          & docs & sents & words & empty & mentions & train \%         & dev \%        \\ \hline
ca\_ancora                                    & 1142 & 12K   & 387K  & 6K    & 56K      & 87.18            & 12.82         \\ \hline
cs\_pcedt                                     & 2212 & 47K   & 1,13M & 34K   & 163K     & 85.08            & 14.92         \\ \hline
cs\_pdt                                       & 2849 & 44K   & 761K  & 19K   & 166K     & 88.10            & 11.90         \\ \hline
cu\_proiel                                    & 19   & 6K    & 61K   & 6K    & 20K      & 86.01            & 13.99         \\ \hline
de\_parcorfull                                & 17   & 1K    & 910K  & 0     & 1K       & 88.73            & 11.27         \\ \hline
de\_potsdamcc                                 & 159  & 2K    & 30K   & 0     & 5K       & 88.77            & 11.23         \\ \hline
en\_gum                                       & 191  & 11K   & 188K  & 100   & 53K      & 87.28            & 12.72         \\ \hline
en\_litbank                                   & 90   & 8K    & 189K  & 0     & 26K      & 88.78            & 11.22         \\ \hline
en\_parcorfull                                & 17   & 1K    & 10K   & 0     & 1K       & 88.36            & 11.64         \\ \hline
es\_ancora                                    & 1211 & 13K   & 420K  & 7K    & 64K      & 88.89            & 11.11         \\ \hline
fr\_democrat                                  & 96   & 11K   & 256K  & 0     & 70K      & 88.97            & 11.03         \\ \hline
grc\_proiel                                   & 14   & 6K    & 65K   & 6K    & 20K      & 93.95            & 6.05          \\ \hline
hbo\_ptnk                                     & 28   & 1K    & 18K   & 0     & 5K       & 42.32            & 57.68         \\ \hline
hu\_korkor                                    & 85   & 1K    & 24K   & 2K    & 4K       & 88.58            & 11.42         \\ \hline
hu\_szegedkoref                               & 360  & 8K    & 117K  & 4K    & 14K      & 89.42            & 10.58         \\ \hline
lt\_lcc                                       & 90   & 2K    & 33K   & 0     & 4K       & 89.89            & 10.11         \\ \hline
no\_bokmaalnarc                               & 315  & 14K   & 225K  & 0     & 68K      & 90.37            & 9.63          \\ \hline
no\_nynorsknarc                               & 364  & 11K   & 191K  & 0     & 57K      & 90.58            & 9.42          \\ \hline
pl\_pcc                                       & 1646 & 32K   & 502K  & 17K   & 169K     & 88.89            & 11.11         \\ \hline
ru\_rucor                                     & 163  & 8K    & 145K  & 0     & 15K      & 85.39            & 14.61         \\ \hline
tr\_itcc                                      & 21   & 4K    & 60K   & 10K   & 19K      & 90.36            & 9.64          \\ \hline
\end{tabular}
\parbox{13cm}{
\captionof{table}{\label{table:corefud-datasets-stats}CorefUD v1.2 statistics~\cite{crac2024}. The statistics for sentences, words, empty, and mentions are replaced with approximate counts by rounding up the numbers.}
}
\end{minipage}
\end{center}




\end{document}